%% file: main.tex
\documentclass{article}
\usepackage{ijcai23}

\usepackage{times}
\usepackage{soul}
\usepackage{url}
\usepackage[hidelinks]{hyperref}
\usepackage[utf8]{inputenc}
\usepackage[small]{caption}
\usepackage{graphicx}
\usepackage{amsmath}
\usepackage{amsthm}
\usepackage{booktabs}
\usepackage{algorithm}
\usepackage{algorithmic}
\usepackage[switch]{lineno}

\usepackage[inline]{trackchanges}
\addeditor{Xu}
\addeditor{Felix}
\addeditor{YJ}
\usepackage{float}
\usepackage{amsmath,amsfonts,amssymb,bm}
\renewcommand{\vec}[1]{\bm{#1}}
\newcommand{\cmark}{\ding{51}} 

\title{Efficient NLP Model Finetuning via Multistage Data Filtering}

\author{
Xu Ouyang\and
Shahina Mohd Azam Ansari\and
Felix Xiaozhu Lin\and
Yangfeng Ji
\affiliations
University of Virginia\\
\emails
\{ftp8nr, dtf8qc, felixlin, yangfeng\}@virginia.edu
}

\input{header.tex}

\begin{document}
\maketitle
\input{abs}
\input{intro}
\vspace{-2mm}
\input{related}

\vspace{-2mm}
\input{method}
\vspace{-2mm}
\input{discussion}
\vspace{-2mm}
\input{eval}

\vspace{-2mm}
\input{summary}
\input{ack}

\bibliographystyle{named}
\bibliography{ijcai23}

\end{document}

%% file: header.tex
\usepackage{wrapfig}
\usepackage{comment}
\usepackage{colortbl}

\usepackage{amsthm} 

\usepackage{color,soul}
\usepackage{graphics}
\usepackage[hidelinks]{hyperref}
\usepackage{lipsum}
\usepackage{tikz}
\usepackage{enumitem}	
\usepackage{listings} 
\usepackage{booktabs}	
\usepackage{lipsum}

\usepackage{capt-of}	

\usepackage{todonotes}  
\usepackage{subcaption} 

\usepackage{mathrsfs}	
\usepackage{amsfonts}

\usepackage[export]{adjustbox} 

\usepackage{algorithm,algorithmic,amsmath} 
\usepackage{boldline}					

\usepackage{multirow}

\definecolor{frenzyorange}{RGB}{249, 158, 26}

\renewcommand{\paragraph}[1]{\vskip 3pt\noindent\textbf{#1 }}	 

  {\begin{list}{$\bullet$}%
     {\setlength{\parsep}{0pt}%
      \setlength{\topsep}{0pt}%
      \setlength{\itemsep}{2pt}}}%
  {\end{list}}
\newcommand\Noted[1]{} 

\newcommand\xzlNote[1]{\sethlcolor{yellow} \hl{#1}} 

\definecolor{darkblue}{rgb}{0.0, 0.0, 0.55}
\definecolor{mygreen}{HTML}{ADFF2F}
\definecolor{mylightgray}{gray}{0.8}

\newenvironment{myitemize}%
  {\begin{itemize}
	[leftmargin=0cm,
		itemindent=.3cm,
		labelwidth=\itemindent,
		labelsep=0pt,
		parsep=1pt,
		topsep=1pt,
		itemsep=1pt,
		align=left]
  }%
  {\end{itemize}}    

  {\begin{enumerate}
	[leftmargin=.cm,itemindent=.5cm,labelwidth=\itemindent,
		labelsep=0pt,
		parsep=1pt,
		topsep=1pt,
		itemsep=3pt,
		align=left]
  }%
  {\end{enumerate}}    


\makeatletter
\def\@copyrightspace{\relax}%
\makeatother

%% file: abs.tex

\begin{abstract}

As model finetuning is central to the modern NLP, we set to maximize its efficiency. 
Motivated by redundancy in training examples and the sheer sizes of pretrained models, we exploit a key opportunity: training only on important data. 
To this end, we set to filter training examples in a streaming fashion, 
in tandem with training the target model. 
Our key techniques are two: (1) automatically determine a training loss threshold for skipping backward training passes; 
(2) run a meta predictor for further skipping forward training passes. 
We integrate the above techniques in a holistic, three-stage training process. 
On a diverse set of benchmarks, our method reduces the required training examples by up to 5.3$\times$ and training time by up to 6.8$\times$, while only seeing minor accuracy degradation. 
Our method is effective even when training one epoch, where each training example is encountered only once. 
It is simple to implement and is compatible with the existing finetuning techniques. 
Code is available at: \href{https://github.com/xo28/efficient-NLP-multistage-training}{https://github.com/xo28/efficient-NLP-multistage-training}
\end{abstract}

\xzlNote{}

%% file: intro.tex

\section{Introduction}
\vspace{-1mm}

\paragraph{Efficient model finetuning}
Modern NLP models are pretrained on large corpora once and then finetuned for specific domains. 
Efficient finetuning is crucial because 
(1) as opposed to one-off pretraining, 
finetuning is invoked for every downstream task and even on individual user's data \cite{houlsby2019parameter}; 
(2) finetuning is often performed close to where the domain training examples reside, e.g. smartphones \cite{rebuffi2018efficient}; 
these platforms often have constrained computing resources. 
As NLP models become larger \cite{devlin2018bert} and language tasks diversify, it is compelling to finetune a model with fewer resources without comprising accuracy much~\cite{zaken2021bitfit,jiang2019smart}.  



Importantly, prior work also has recognized the importance of efficient finetuning. 
A popular approach is to impose efficient model structures~\cite{sun2019fine}, including low rank approximation \cite{lan2019albert,ma2019tensorized}, weight sharing \cite{dehghani2018universal,lan2019albert}, knowledge distillation \cite{hinton2015distilling}, pruning \cite{cui2019fine,mccarley2019structured}, quantization \cite{jacob2018quantization}, and layer freezing \cite{lee2019would}.

While most of the work focuses on modeling strategies, this paper exploits an opportunity particularly for finetuning with large datasets:
\textit{filtering training data at low computational cost}. 
Given that training data is known as often redundant \cite{katharopoulos2018not},
we test a simple idea: skip training examples that are less important to the gradient updates. 

\input{tab-overview}

\paragraph{Challenges}
The idea raises twofold challenges. 
First, how to assess training example importance? 
As the model is being updated throughout a training process, 
the assessment must weigh training examples against the model's \textit{up-to-date} capability. 
Second and more importantly, 
the assessment itself should incur low computational overhead. 
To this end, much prior work for selecting training data does not apply \cite{mirzasoleiman2020coresets,wang2020optimizing}: 
targeting training effectiveness but not efficiency, they often use computationally expensive methods to weigh data, e.g. comparing training gradients; 
they often slow down training or incur costly prepossessing before training. 



\paragraph{Our method}
We use training loss as the signal for data importance: 
low loss means the model has high confidence in the training examples, which could be skipped to avoid the cost.
As will be shown in the paper, this simple idea is powerful:  
skipping training data on which losses are lower than a fixed, hand-picked threshold $L_{low}$ can skip 20\% -- 50\% of the data 
while seeing a minor ($<$1\%) drop in the final model accuracy. 
Yet, hand-picking $L_{low}$ is tedious; 
comparing $L_{low}$ against losses still requires the computation of forward passes. 



To this end, we propose an algorithm that learns to
predict data filtering decisions in tandem with training. 
Given a model and training data, the algorithm automatically derives a proper loss threshold $L_{low}$ and further skips forward and/or backward passes on training data, whenever appropriate.

The algorithm runs training as a multistage process: each stage receives supervision from earlier stages but is more efficient than the former. 
After training starts: 
\begin{myitemize}
\item 
The first stage derives $L_{low}$ that adapts to the model and the training data. 
This stage runs both forward and backward passes on training examples. 

\item 
The second stage uses the derived $L_{low}$ to filter backward passes.
With the filtering decisions, it trains a meta predictor based on simple linguistic features
of input texts. This meta predictor will later decide if the given examples are worth training.

\item 
The third stage queries the meta predictor to filter both forward and backward passes. 
\end{myitemize}

The first two stages are short (processing average 16.1\% of
examples across all benchmarks) while the most efficient third stage processes most examples.
The algorithm automatically advances across the stages, based on its observation of training losses and the meta predictor's performance. 

\paragraph{Results}
On a diverse set of NLP benchmarks, 
our algorithm reduces the total training time by up to 6.7$\times$ (5.88$\times$ on average). 
The resultant accuracy degradation is minor, no more than 1.44\% (0.6\% on average). 
An ablation study shows the efficacy of our techniques: 
(1) the automatic loss threshold skips backward passes for up to 84\% of the training examples; 
(2) the meta predictor skips forward passes for up to 81\% training examples when trained for 2 epochs;
(3) as the number of epochs grows, our efficiency is increasingly higher, e.g. up to 18.1$\times$ training time reduction with 1.83\% lower accuracy when training for 5 epochs on SST2 \cite{wang2019glue}.

\paragraph{Contributions}
\begin{myitemize}
\item 
We presented empirical evidence that large NLP datasets
are redundant for model finetuning, and many training examples can be filtered with a low impact on accuracy. 

\item 
We proposed a simple, automatic mechanism for filtering training data:
using an automatic loss threshold for skipping backward passes and a lightweight meta predictor for further skipping forward passes.

\item 
We presented a holistic training process that integrates the above techniques and demonstrated its efficacy on diverse NLP text classification tasks. 

\item
We proposed a novel, comprehensive evaluation strategy: using a new metric considering both the accuracy and efficiency; estimating the energy consumption to reduce CO$_2$ emission.
\end{myitemize}

%% file: tab-overview.tex
\begin{table}[t]
		\centering
	\includegraphics[width=0.48\textwidth]{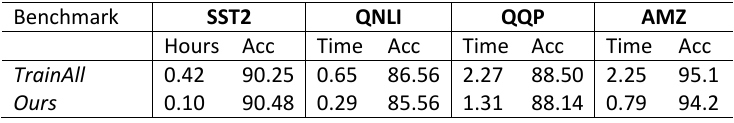}
		\vspace{-6mm}
	\caption{Our method significantly reduces training time (in hours) while achieving accuracies comparable to training with all the data (\textit{TrainAll}). Details in \S\ref{sec:eval}.
	GPU: Nvidia RTX 2080Ti.}
	\label{tab:highlight-res}
		\vspace{-4mm}
\end{table}

%% file: related.tex

\section{Related Work}
\vspace{-1mm}
%
\paragraph{Effective training}
Aggressive-passive training (APT) \cite{shalev2003online} and Perceptron \cite{rosenblatt1958perceptron} are well-known online learning algorithms that only update the model either on high-loss or misclassified samples. 
Unlike our method that can skip \textit{both} forward and backward passes, these methods would require forward passes on all the training data, which is much less efficient.

Curriculum learning \cite{bengio2009curriculum} trains a model from the easiest examples to the hardest ones, hence accelerating model convergence. 
Yet, arranging the data order requires prior domain knowledge of all training data, e.g. how ``noisy'' are these examples, which can be very expensive to preprocess a large dataset to collect such information.
To get rid of this requirement, self-paced learning \cite{NIPS2010_e57c6b95}
updates model weights by considering the level of difficulty of given examples. 
Unlike us, these techniques do not filter training data; 
they incur high optimization costs on large NLP models and datasets.

\paragraph{Training data selection}
Motivated by training data redundancy, previous work downselects the data; 
however, many require expensive data preprocessing, making them expensive for NLP finetuning. 
Importance sampling~\cite{katharopoulos2018not} can train with smaller data batches that have a similar gradient norm as the full data batches. 
CRAIG~\cite{mirzasoleiman2020coresets} selects training examples for which the weighted sum of gradients closely approximates that of the full training set. However, these two methods sample important data by solving the optimization problem for each data to mimic the \textit{full} data gradient. 

\textbf{Filtering training data with loss} is known. 
Yet, our contributions are (1) a meta predictor for deciding filtering and (2) a multistage process for training the meta predictor. Hence, our proposed training procedure can automatically skip forward and backward computations, a goal unattainable for loss-based filtering. \S\ref{sec:eval} shows that we outperform loss-based filtering.

Our goal is related to active learning but not the same: 
in an unsupervised setting, 
it down-selects unlabeled data for labeling and then training~\cite{settles2009active}. 
However, few active learning methods that employ a multistage process, 
which first trains a meta predictor in a \textit{supervised} fashion and then uses the predictor to filter data in an \textit{unsupervised} fashion. 
To this end, our idea can be extended to support active learning. 

Clustering training data can remove at least 10\% examples for object recognition \cite{birodkar2019semantic}. But it needs to sweep through all the data instances and keep updating clusters before picking out close ones. 
Selection via proxy \cite{coleman2019selection} simplifies the big target model to a small proxy model finding the core-set to train the big target model. Unlike us, the proxy model needs to be trained on all the data samples then they train the target model separately.

AutoAssist~\cite{zhang2019autoassist} shares our motivation: to filter  training data with a small model called ``assistant''. 
Although the paper shows that AutoAssist reduces the \textit{training loss} at a higher rate, it was unclear by how much \textit{training time} is reduced. 
The assistant is complex, e.g. stochastic sampling data to generate a batch, which takes more than ten epochs to warm up, unsuitable to NLP finetuning which comprises no more than several epochs.
By contrast, our design is simple, saving training time significantly even for one epoch. The saving is increasingly higher as the epoch count grows. 
We will experimentally compare to AutoAssit in \S\ref{sec:eval}.

\paragraph{Other training optimizations}
include accelerating model convergence with the same amount of data, e.g. 
by varying the learning rate per weight \cite{jacobs1988increased,zeiler2012adadelta} or batch sizes \cite{smith2017don}.
There also exist techniques for reducing training computation by reducing model parameters or finding a lightweight counterpart of a large model: 
pruning \cite{cui2019fine,mccarley2019structured,katharopoulos2018not},
knowledge distillation (KD) \cite{hinton2015distilling,sanh2019distilbert,jiao2019tinybert},
quantization \cite{jacob2018quantization,micikevicius2017mixed},
low-rank approximation \cite{lan2019albert,ma2019tensorized}, and weights sharing \cite{dehghani2018universal,lan2019albert}. 
Our method is complementary to these optimizations and can be used in conjunction with them. 

%
%
%

%% file: method.tex
\section{Our Method}
Let us assume that the whole training set is divided into $M$ mini batches $\mathcal{D}=\{D_m\}_{m=1}^M$ and each mini batch consists of $N$ training examples.
The training loss for the $m$-th mini batch is defined as:
\vspace{-1mm}
\begin{equation}
\label{eq:loss}
  l(m) = \frac{1}{N}\sum_{n=1}^{N}CrossEntropy(\hat{y}^m_n, y^m_n)
\end{equation}
where $\hat{y}^m_n$ is the model prediction and $y^m_n$ is the ground truth.
Intuitively, if the model is confident with all the examples in this mini batch, we should expect that the average loss $l(m)$ will be very small, and skipping the training on this mini batch would not cause a big difference in model performance. The strategy of skipping individual examples allows for a more granular level of impact control.
Furthermore, we hypothesize that, for a given input sentence, the words in the sentence may provide enough clues about the model's confidence level.
For example, if the model gives high confidence prediction on a group of words, we should expect it to show high confidence in sentences having the same words.

We address the following design questions: 
\begin{myitemize}
\item How to select mini batches and skip their backward passes, given their training loss? 
\item Furthermore, how to select mini batches and skip their forward passes, 
\textit{without} calculating their training losses?
\item How to realize the two above mechanisms in an efficient, automatic manner?
\end{myitemize}

\vspace{-1mm}
\subsection{Three-stage training}
\label{3stages}

\input{fig-workflow}

Our proposed algorithm consists of three consecutive stages; 
the dataflow of each stage is shown in Figure~\ref{fig:workflow}.
Throughout the process: stage 0 warms up training and estimates a loss threshold for selecting mini batches; 
stage 1 filters out mini batches by referring to the estimated threshold, meanwhile using the filtering results to train a meta predictor; 
stage 2 uses the meta predictor to filter out mini batches without explicitly calculating their forward losses. 
Each stage requires supervision in earlier stages to become effective but is more efficient than the former. 

\paragraph{Stage 0: estimating loss threshold}
This stage warms up the model training and automatically derives the loss threshold $L_{low}$.
For each mini batch, the algorithm runs both the forward step for computing the training loss and the backward step for gradient update. 
After training with the initial several mini batches, the loss threshold is estimated by the moving average of recent training losses:
\vspace{-1mm}
\begin{equation}
 L_{low}(m) = \frac{1}{K}\sum_{k=1}^{K} l(m-k)
\end{equation}
where $l(m)$ is the training loss calculated on the $m$-th mini batch defined in \autoref{eq:loss}, and $K$ is the window size of moving average. 
Prior work already used the moving average of training losses to estimate the trend of loss changes and monitor the training process \cite{zhang2019autoassist}.
This work further uses the moving average to identify mini batches worth training, i.e. if the training loss on a mini batch is higher than the moving average. 
Our study shows that, with a small window ($K=8$), 
the loss threshold quickly stabilizes with low variations. 
For instance, the variation becomes less than $4.3\times10^{-5}$ on QNLI when stage 0 spans the first $7.7\%$ of training data. 

\paragraph{Stage 1: training the meta predictor}
With a stabilized loss threshold $L_{low}$, the algorithm moves to the first stage of training, in which it can filter out some mini batches if their losses are lower than $L_{low}$. 
To further prepare for stage 2, the algorithm starts to build a meta predictor $f$, which aims to predict the filtering decision for a mini batch \emph{without} resorting to the forward pass.
Specifically, we implement the meta predictor $f$ as a binary naive Bayes classifier:
\begin{equation}
\label{eq:naive-bayes}
f(\vec{w}^m)\in\{0, 1\}
\end{equation}
where $\vec{w}^m$ represents the bag-of-words representations of the input sentences in the $m$-th mini batch; 
the predicted value 1 indicates that the training loss is likely to be higher than the threshold, while 0 indicates that the loss is likely to be lower than the threshold. 
With the forward computation on mini batch $m$ and the loss threshold $L_{low}$, this mini batch provides a training example for updating the meta predictor $f$.

In order to measure how much information the meta predictor learns, 
we use the same notations as above and additionally define $y'$ as the ground truth label, 
which indicates if elements in the mini batch $m$ are worth training by comparing their forward losses with $L_{low}$. 
The meta predictor's loss is defined as: 
\vspace{-1mm}
\begin{equation}
\label{eq:predictorloss}
  l_{mp}(m) = -\frac{1}{N}\sum_{n=1}^{N}\log p({y'}^m_n\mid \vec{w}^m_n)
\end{equation}
If the meta predictor exhibits sufficient low moving average losses, stage 1 ends.

In a nutshell, the meta predictor further saves the computational cost if it deems that the target language model is likely familiar with the examples in a mini batch. 
For this purpose, it is sufficient for the predictor to run Naive Bayes classification on the input words because lexical information is an essential component of a text's semantic information.
The following section on~\autoref{sec:discussion} will present more design choices. 

\paragraph{Stage 2: filtering data with the meta predictor}
Once the meta predictor shows adequate performance, 
the algorithm queries it for screening each mini-batch. 
On a mini-batch that the predictor deems worth training, the algorithm runs a forward pass and uses the resultant loss to decide if a backward pass is needed. 
Hence, stage 2 can skips both forward and backward passes for high efficiency.

With our design, the training process reaches stage 2 soon after the first epoch starts, as we will demonstrate in the evaluation.
In case the training spans more than one epoch, the process remains in stage 2 in subsequent epochs.

\vspace{-1mm}
\subsection{A Special Case}


While the three-stage algorithm is effective (as \autoref{sec:eval} will show), it is possible to use a stripped-down version of it: 
after determining $L_{low}$ automatically, use it to filter all the remaining training data without invoking the meta predictor. 
Doing so would miss the opportunity of skipping forward passes; 
yet, by eschewing the meta predictor and its hyperparameter tuning, the method further simplifies training. 
We refer to this method as \textit{automatic threshold only} and will compare to it in \autoref{sec:eval}.


%% file: fig-workflow.tex

\begin{figure}[t]
\centering
\includegraphics[scale=0.4]{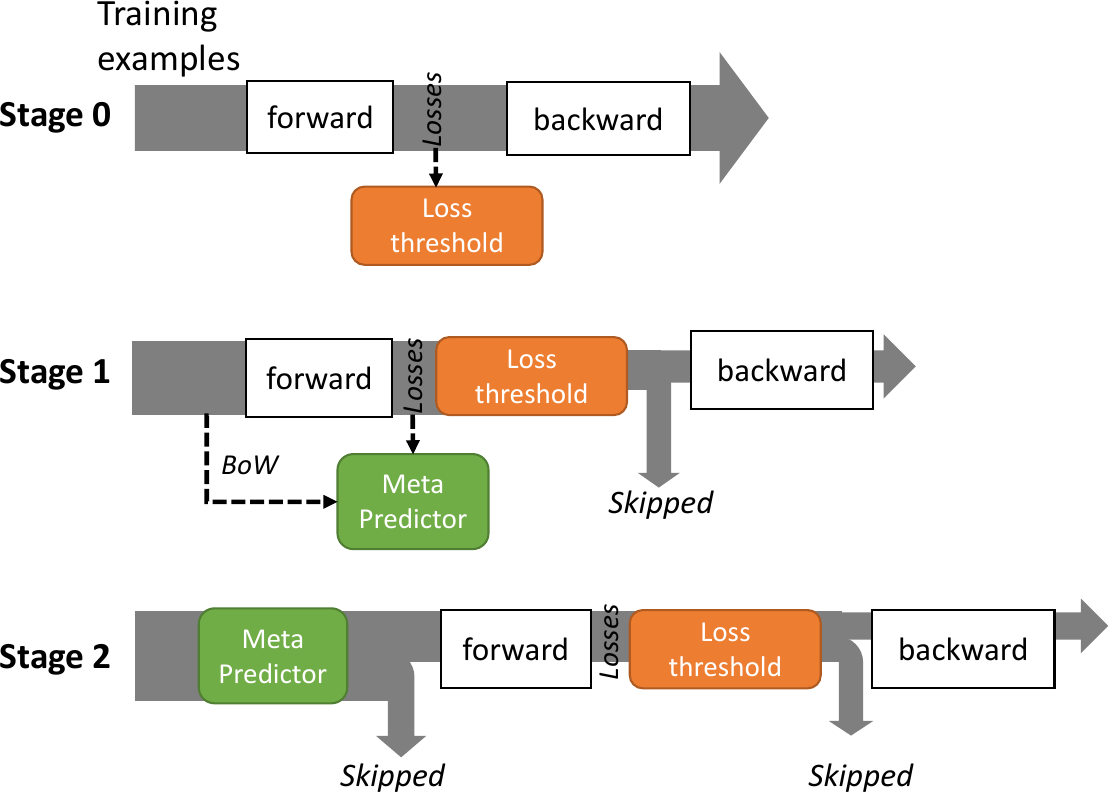}
\caption{An overview of our three-stage training algorithm}
\label{fig:workflow}
\vspace{-3mm}
\end{figure}

%% file: discussion.tex
\section{Discussion on Design Choices}

\label{sec:discussion}

\subsection{Automatic loss thresholds}
\vspace{-1mm}

\paragraph{Loss for filtering data}
We use losses for the following reasons. 
(1) 
As shown in prior work~\cite{loshchilov2015online}, 
losses effectively reflect model update from given examples: 
higher loss means that the model makes higher inference errors and hence can learn more from this example. 
(2) 
Getting losses is computationally inexpensive: they incur no computation overhead beyond forward passes. 

\paragraph{Rationales for a loss threshold} 
We want to selectively learn from a subset of examples with the highest losses. 
To identify such a subset, one may attempt to compare the losses of all training examples.
This however suffers from two drawbacks.
(1) It is inefficient: doing so requires forward passes on \textit{all} the training data, an expensive task with poor data locality, 
because the activations calculated by forward passes must be saved to disk and later restored for executing backward passes. 
(2) As a model is being updated, the losses of examples yet to be trained will  change. 
For instance, a previously high-loss example (estimated by the untrained model) may see a low loss and hence is no longer worth training.


By contrast, we use a loss threshold to screen examples. 
(1) The method is effective because it accommodates continuous model updates. 
As the loss is always assessed based on the updated model, the method estimates how much the \textit{updated} model can learn from the example under question. 
(2) The method is efficient because it consumes training data sequentially with good locality; it is, therefore, friendly to memory hardware hierarchy. 
(3) The method leads to self-paced training. 
As training starts, the model likely sees high losses on most data, for which it will filter less; as training goes on and the model is updated, it likely sees lower losses on more data, for which it will filter more. 

\paragraph{Tradeoffs for setting a loss threshold} 
Ideally, we want to train on the fewest examples and have an accuracy close to that from the \textit{TrainAll} baseline. 

Hand-picking $L_{low}$ is tricky: 
a $L_{low}$ too high can be overly \textit{passive}~\cite{shalev2003online}, 
resulting in too much data filtered and suboptimal accuracy; 
a $L_{low}$ too low can be overlay \textit{aggressive}, resulting in long training delays. 
The optimal choice of $L_{low}$ hinges on a discrepancy between the model's knowledge prior to training and the total knowledge encoded in the training data. 
Unfortunately, this discrepancy cannot be accurately determined until we have trained on all the data. 
This motivates us to derive $L_{low}$ automatically.

\paragraph{Automatic loss thresholds}
We set $L_{low}$ to be the average of the most recent losses for the following reasons.
(1) By considering a sample of the total training set, the average loss estimates how the current model fits the data yet to be trained with; 
(2) By picking $L_{low}$ to be the \textit{average} loss, we balance being passive and being aggressive in filtering examples. 
(3) We keep updating $L_{low}$ in sliding windows so that we keep refreshing the estimation.
\vspace{-1mm}
\subsection{The loss predictor}
\vspace{-1mm}


Our meta predictor addresses two design questions. 
First, what features should the predictor be based on?
That extraction of such features must be significantly cheaper than a forward pass of the language model $M$ under training. 
We adopt Bag-of-Words (BoW) features of an input sequence as input of the predictor. Bag-of-words is one of the classical text features for different NLP models and tasks~\cite{sebastiani2002machine,heckerman1997bayesian,lewis1998naive}.
Apart from BoW features, other text classification features can also be used.

Second, who is responsible for training the meta predictor, which shall be specific to the model $M$ and the training data? 
We train the predictor under the supervision of loss-based example filtering (stage 1). 
The training does not have to be long before the predictor can be queried to make predictions and advise on filtering training data.

It is worth noting that even when the trainer queries the predictor for  filtering decisions (stage 2), it still updates the predictor continually.  
This is done on the data that the predictor deems worth training: 
the trainer runs forward passes on such data, get the losses, and uses the comparison outcome to update the predictor. 
This keeps the predictor updated to the changing language model $M$: 
as $M$ is being trained, the correlation between its loss on new data and the data's BoW features is drifting; 
intuitively, $M$ will see a lower loss given the same BoW features. 

The cost of meta predictor training (naive Bayes) is two orders of magnitude lower than the target model (DistillBERT). 
%
As stated in \S\ref{3stages}, training stops when the meta predictor’s average loss drops below a threshold, ALT. Section \S\ref{hyperparameters} further tests a range of ALT values.

%% file: eval.tex

\section{Evaluation}
\label{sec:eval}
\vspace{-1mm}
We set to answer the following questions: 
\begin{myitemize}
\item Compared to the existing training method, 
can we achieve comparable accuracy with much higher efficiency? 
\item How significant are our key techniques?
\item How sensitive is our method to its hyperparameters and what are their reasonable ranges?
\end{myitemize}

\vspace{-1mm}
\subsection{Experiment Setup}
\vspace{-1mm}

\paragraph{Models \& Datasets}
We test our method on the pretrained DistilBERT model with 6 transformer layers and a hidden dimension of 768. 
We finetune it on five classification benchmarks.
Three are from GLUE \cite{wang2019glue},
one is the Amazon Polarity (AMZ) \cite{zhang2015character}, and one multilabel benchmark is AG News (AG) \cite{Zhang2015CharacterlevelCN}
The benchmarks cover tasks of single-sentence, similarity, and inference. 
All the benchmarks have substantial training data, allowing opportunities for filtering. 
For the three GLUE benchmarks, we reproduce the accuracies reported in prior work \cite{sanh2019distilbert} and consider them as the baseline performance; 
for Amazon Polarity and AG\_news, we finetune the model and report accuracy measured on its test set.
We choose these datasets by following DistillBERT's experiment plan and they are substantially large. Our methods only rely on losses and semantic information. Therefore, they can be easily applied to various NLP tasks by loading different pretrain models.
The comparable efficacy of batch and example skipping prompted our inclusion of results from the latter approach.

\paragraph{Baselines} 
We evaluate against four groups of baselines. 
(1) \textit{TrainAll} trains models with all the training data.
(2) \textit{FixedThreshold} filters the training data with hand-picked, fixed loss thresholds, for which we sweep the range $[0.1,0.7]$ at an increment of 0.2. 
(3) \textit{AutoAssist}~\cite{zhang2019autoassist} filters out instances with a lightweight ``assistant'' model jointly with the target ``Boss'' model. The assistant selects and generates batches during training both models. 
(4) \textit{Selection via Proxy} (SVP)~\cite{coleman2019selection} uses a small proxy model to perform core-set selection and train the target model with the core-set.




\input{tab-hyperparam}

\paragraph{Metrics}
For a training process, we report (1) the model accuracy after training and (2) the training time, which is inverse to training efficiency. We make experiment results reproducible. 
On our machine with Nvidia RTX 2080 Ti, 
we measure the time of a forward pass ($T_f$) and a backward pass ($T_b$), as well as the fractions of data on which only backward passes are skipped ($\alpha_b$) and \textit{both} forward and backward passes are skipped ($\alpha_{fb}$) in each training process. 
For a given training process, the total time is $T=\alpha_b T_f + (1-\alpha_b-\alpha_{fb}) (T_f + T_b)$. 
Finally, we normalize $T$ to that of \textit{TrainAll} as
$T_{norm}= \frac{T_{ours}}{T_{all}}$ and report $T_{norm}$.

\input{fig-pareto}

\input{tab-e2e-combined}

\vspace{-1mm}
\subsection{End-to-end results}
\autoref{fig:pareto} plots accuracy versus training time. 
It shows that our method delivers both high accuracy and high efficiency (i.e. low training time). 
Compared to \textit{TrainAll}, our method reduces the training time by at least 2$\times$. 
Meanwhile, the accuracy is similar: 
on QNLI, our best accuracy is only about 1\% lower than that of \textit{TrainAll};
on QQP, AMZ, and AG, our accuracy is within 1\% of that of \textit{TrainAll}. 

Note that on SST2 our accuracy is even higher than \textit{TrainAll}, 
which we attribute to that our method excludes training examples that the model is already confident about, therefore preventing overfitting on such examples. 



Our method provides rich trade-offs between training accuracy and efficiency. 
\autoref{fig:pareto} highlights a series of \textit{Pareto} results as desirable trade-offs~\cite{deb1999multi}: 
given a result, no other result has both higher accuracy and higher efficiency. 


\paragraph{Accuracy gain over time (AGOT).}
Prior efficient training research often compares accuracies under a fixed computation budget and vice versa. 
Yet, using a single evaluation metric can be inadequate, as neither accuracy or runtime could characterize the tradeoffs between them in a comprehensive fashion.
To this end, a user may want to quantify her most desirable accuracy/efficiency tradeoff. 
We therefore define AGOT: 
\vspace{-1mm}
\begin{equation}
\label{eq:agot}
\text{AGOT}_{\varepsilon}(a, t) = \frac{a - a_{base}}{a_{full} - a_{base}}\cdot \frac{1}{T_{norm}^{1-\varepsilon}}
\end{equation}
where $a$ is the model accuracy after training and $T_{norm}$ is the training time normalized to \textit{TrainAll}; 
$a_{base}$ and $a_{full}$ are model accuracies before training and after training with all the data, respectively. 
In the above definition, the accuracy gain (the first term) is inversely weighted by the running time needed to achieve this gain (the second term). 
The parameter $\varepsilon\in[0,1]$ is decided by the user, 
reflecting her preferred importance of accuracy with regard to the training time. 
A larger $\varepsilon$ weighs more on the accuracy; 
a special case $\varepsilon=1$ means not considering the time at all.
Specifically, we set $\varepsilon=0.95$, weighting significantly on the accuracy: 
a few points of accuracy gain often warrants significantly longer training time as shown in \autoref{fig:pareto}. 

To compare against baselines, for each benchmark we consider the result with the highest AGOT score, referred to as AGOT-optimal. 
As shown in \autoref{fig:pareto}, AGOT effectively identifies accuracy/efficiency sweet spots. 
On QNLI and QQP, the AGOT-optimal results are the ones with the highest accuracy among all the results with slightly longer training time than the latter.
On AMZ and AG, the AGOT-optimal result has slightly lower accuracy 
but much lower training time. 
The AGOT-optimal results are highly competitive against \textit{TrainAll}. 
Take SST2 as an example, the AGOT-optimal reduces the training time by 85\% while showing superior accuracy by 0.69\%, likely because noisy training data is filtered. 

We next focus on AGOT-optimal results. 


%

\paragraph{How much computation is skipped?}
Our method skips large fractions of forward and backward passes. 
As \autoref{tab:e2e-combined} (b) shows, across all benchmarks, on 52.87\% -- 81.01\% of the training data both forward and backward passes are skipped. 
This indicates that our predictor, which controls skipping both forward and backward passes, is highly effective. 
In addition, on 8.95\% of the training data on average, backward passes are skipped while forward passes are executed. 

\paragraph{Comparison versus prior works}
The results are shown in Table~\ref{tab:e2e-combined}.
Compared to \textit{AutoAssist}, our method trains for much less time (up to 5.33$\times$) while achieving similar/higher (-0.13\% -- +0.46\%) accuracy.
AutoAssist's disadvantage is likely because its choices of filtered data is sub-optimal:
a large fraction of training batches are generated via random selection with replacement,
which forces the simple assistant model to learn duplicate samples. 
Furthermore, since the assistant is being updated throughout the whole training process,
overfitting is likely to occur.


We evaluate \textit{SVP} on AMZ and AG (not in Table~\ref{tab:e2e-combined}),
as SVP's data preparation code is incompatible with the remaining benchmarks. 
On the benchmarks, SVP's accuracies are 88.54 and 90.02, respectively; 
the normalized training times are both 0.4.
Compared to SVP, our method's accuracies are 5.54\% and 3.62\% higher and our training time is  1.82$\times$ and 2.67$\times$ shorter. 
This is because SVP's core-set selection depends on the consensus of data valuation between the proxy and the target models, which does not always hold. 

It is worth noting that 
both \textit{AutoAssist} and \textit{SVP} need to train significantly more epochs than ours. 
For example, AutoAssist needs 100 epochs on image and language tasks. 

\paragraph{Estimated energy \& CO$_2$ reduction}
We use an energy model~\cite{strubell2019energy}: 
\vspace{-1mm}
\begin{equation}
\begin{split}
&p_t = \frac{1.58t(p_c+p_r+gp_g)}{1000}; CO_2e = 0.954p_t
\end{split}
\end{equation}

In the equation, $p_t$ is the total energy consumed during finetuning, $p_c$ is the average CPU power draw, $p_r$ is the average DRAM power draw, $p_g$ is the average GPU power draw, $t$ is the training time, and $g$ is the GPU count. 
Compared to \textit{TrailAll}, we reduce the total energy consumption by 45.85\% on average across all benchmarks. 
Considering the proportions of different energy sources in the US~\cite{strubell2019energy}, 
we estimate to reduce the CO$_2$ emission from 1.05 pounds per training process to 0.56 pounds on average.


\input{tab-epochs}



\input{tab-random}

\vspace{-1mm}
\subsection{Ablation Study}
\vspace{-1mm}


\paragraph{Efficacy of using loss thresholds}
The results are shown in Table~\ref{tab:cmp-random}. 
Compared to filtering the same amount of training data that is randomly selected,
methods based on loss thresholds show consistently higher accuracies. 
Specifically, \textit{FixThreshold} with loss thresholds of $[0.1, 0.3, 0.5]$ (the skip ratio varies from 20.72\% to 83.25\%) shows accuracies higher by 0.93\% -- 1.33\% on average.
The Three-stage method shows accuracies higher by 
2.14\% on average.
Note that such accuracy improvement is significant:
take Figure~\ref{fig:pareto} as reference, one can reduce the training time by 5$\times$ with tolerating accuracy drop as low as 2.01\% by average.


\paragraph{Efficacy of automatic threshold}
\textit{AutoThreshold} can find a loss threshold that results in competitive accuracy and efficiency. 
\autoref{tab:e2e-combined} compares \textit{AutoThreshold} to \textit{FixedThreshold}, showing that  
the former delivers higher AGOT than all the fixed thresholds tested; 
in fact, it delivers \textit{both} higher accuracy and lower training time than most of the fixed thresholds.


\paragraph{Efficacy of the meta predictor}
The meta predictor is essential to our efficiency as it skips a large fraction of forward passes, as shown in \autoref{tab:e2e-combined}. 
Compared to \textit{AutoThreshold} which can only skip backward passes, 
our Three-stage training reduces the training time by \textit{additional} 2.01$\times$ on average across all benchmarks.
Furthermore, the accuracy is higher by 0.57\% on average, which is likely because the meta predictor better learns training data importance as the training proceeds. By our design, the data filtered by the loss threshold and meta predictor should be highly overlapped, which is shown that up to 70\% of the filtered data overlapped.


\vspace{-1mm}
\subsection{Sensitivity to hyperparameters}
\label{hyperparameters}
\autoref{tab:hyperparam} summarizes our hyperparameters.
We next study their impact on SST2 performance and their reasonable ranges. 



\paragraph{Number of epochs ($N_{epoch}$)}
As shown in \autoref{tab:epochs}, our efficiency will be more pronounced as $N_{epoch}$ increases. 
Overall, the first epoch gains most of the model accuracy; 
additional epochs yield diminishing return or even fluctuation. 
This is consistent with prior observations \cite{sanh2019distilbert,jiao2019tinybert} and is the reason why users commonly finetune an NLP model for no more than several epochs.
As $N_{epoch}$ increases, our method has similar accuracy (within $\pm 0.6\%$) as \textit{TrainAll} while seeing increasingly higher reduction in the training time,  e.g. 2.36$\times$ when $N_{epoch}$ = 1 and 6.53$\times$ when $N_{epoch}$ = 5. 
This is because only in the first epoch our method runs stage 0 and 1, paying the learning cost; in subsequent epochs, our method remains in stage 2,  invoking the meta predictor to skip most of the forward and backward passes. 




\paragraph{Number of mini-batches ($N_0$)} 
The fraction of data examples in stage 0. We try fraction of 10\% -- 40\%. In this stage, the model does forward and backward passes on all the data.

Accuracies are stable with the increase of $N_0$. From 10\% -- 40\%, average accuracy is 89.68\% $\pm$ 0.35\%.
But normalized run time grows along with $N_0$. $N_0$ = 40\% has 1.95$\times$ longer runtime than $N_0$ = 10\%.
Experiment results show AGOTs are not sensitive to $N_0$, as the differences among them are very small.
$N_0$ = 30\% has the lowest AGOT because of lowest accuracy, however $N_0$ = 40\%'s highest accuracy compensates its longest run time. 
When Sliding window size ($W$) and Average loss threshold ($ALT$) = (8 or 16,  0.1 or 0.2), training will not reach stage 2.
This because it's hard for the meta predictor to be trained as well as required by these harsh conditions. But it also proves $N_0$ is not a hyperparameter controlling filtering.

\paragraph{Sliding window size ($W$)} 
The sliding window size in stage 1 for collecting predictor losses. The stage transition is determined by the average loss in this window.

From the results, both accuracies and normalized run time grow with $W$. The average accuracy for $W$ = $[4, 8, 16]$ is 89.74\% $\pm$ 0.29\%
and average run time is reduced by 4.77$\times$ $\pm$ 0.64\%.
Based on this trend, average AGOTs are very stable. The highest AGOT difference ratio is only 0.11\%. The training time increment is because a larger $W$ has higher demand on the meta predictor and it takes longer time in stage 1 and shorter time in stage 2. 
When $W$ and $ALT$ = (8 or 16, 0.1 or 0.2), training will not switch from stage 1 to stage 2.
We can find that the switching likely fails under bigger $W$ and lower $ALT$, because they are more difficult condition to fulfill. So $W$ actually controls data filtering.

\paragraph{Average loss threshold ($ALT$)}
The average loss threshold in stage 1 measuring whether the predictor has been well trained. As long as the average loss in $W$ is lower than $ALT$, training switches to stage 2.

With the increment of $ALT$, accuracies decrease. The average accuracy of $ALT$ = $[0.1, 0.2, 0.3, 0.4, 0.5]$ is 89.37\% $\pm$ 0.36\%,
their differences are between 0.06\% -- 0.89\%. The normalized run time decreases with $ALT$ from 4.39$\times$ to 6.49$\times$ of \textit{TrainAll}'s run time. Average AGOTs are also stable, its maximum difference is only 1.2\%. 
Similar with $W$, $ALT$ controls whether training goes to the last stage or how much filtering we do.

\input{tab-freeze}
\vspace{-2mm}
\subsection{Compatibility with layer freezing (LF)}
\vspace{-1mm}
Our method complements LF (i.e. training only last few layers), a common optimization for finetuning~\cite{sun2019fine,lee2019would} . 
Table~\ref{tab:freeze} compares \textit{TrainAll} and our method both freezing all but the last layer. 
First, our method is compatible with LF. 
Compared to \textit{TrainAll} with LF, our method with LF achieves comparable accuracy (lower by 0.46\% -- 1.61\%) in much lower training time (lower by 1.75$\times$ -- 7.14$\times$). 
Second, our method is still relevant when LF is in use: applying LF to \textit{TrainAll} reduces the training time by 2.34$\times$ with accuracy loss of 2.88\% on average; 
by comparison, our method can reduce additional 2.78$\times$ lower training time at 3.29\% accuracy loss. 
The results encourage use of our method and LF in conjunction. 



%% file: tab-hyperparam.tex

\begin{table}
\centering
	\includegraphics[width=0.48\textwidth{}]{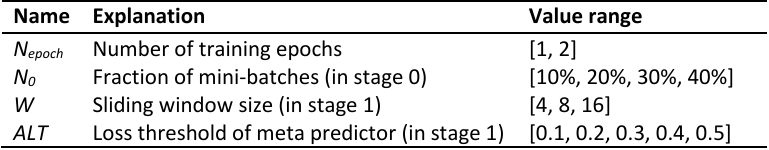}
	\vspace{-1em}
	\caption{A summary of hyperparameters. 
	The value range column shows concrete values used in evaluation.}
	\label{tab:hyperparam}
 \vspace{-3mm}
\end{table}

%% file: fig-pareto.tex

\begin{figure*}[t]
	\centering
	\includegraphics[width=0.95\textwidth]{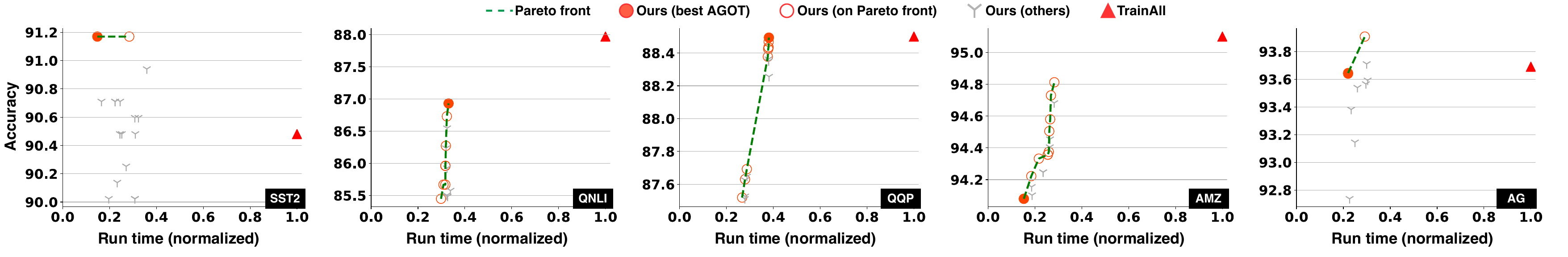}
		\vspace*{-2mm}
	\caption{Compared to training with all data, our method achieves comparable accuracy in a much shorter training time (i.e. higher efficiency). Each plot shows multiple results of our method, resulted from different hyperparameters. 
	}
	\label{fig:pareto}
\end{figure*}

%% file: tab-e2e-combined.tex
\begin{table*}[t]
	\centering
	\includegraphics[width=0.95\textwidth{}]{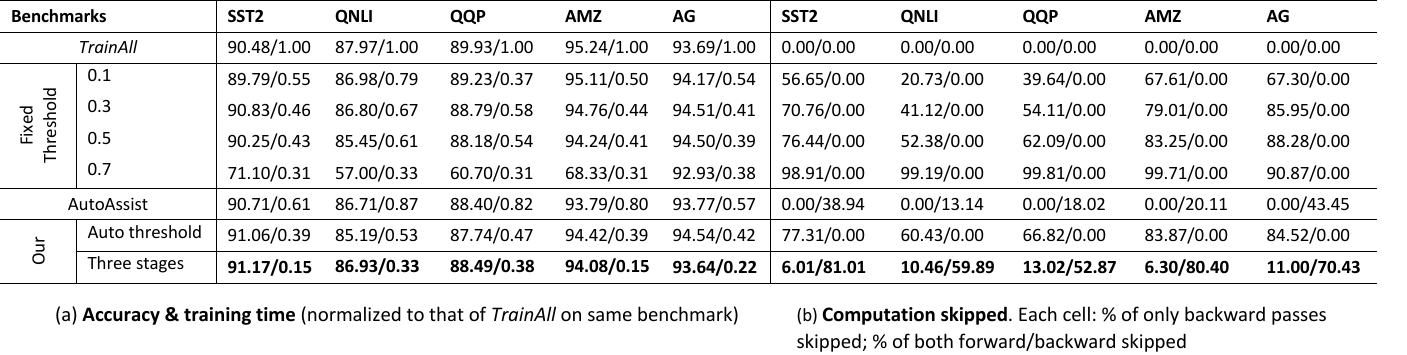}
	\vspace{-2mm}
	\caption{Our method as compared to the baselines. For our method, only the AGOT-optimal results are shown.}
	\label{tab:e2e-combined}
	\vspace{-3mm}
\end{table*}

%% file: tab-epochs.tex
\begin{table}[t]
	\centering
	\includegraphics[width=0.45\textwidth{}]{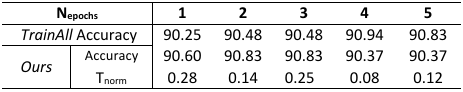}
	\vspace{-2mm}
	\caption{
	Accuracy and training time ($T_{norm}$) as the epoch count ($N_{epochs}$) grows, showing that our method yields increasingly higher efficiency in additional epochs. 
	$T_{norm}$ is normalized to the time of \textit{TrainAll} of the same $N_{epochs}$. 
	}
	\label{tab:epochs}
\vspace{-3mm}
\end{table}

%% file: tab-random.tex

\begin{table}[t]
	\centering
	\includegraphics[width=0.3\textwidth]{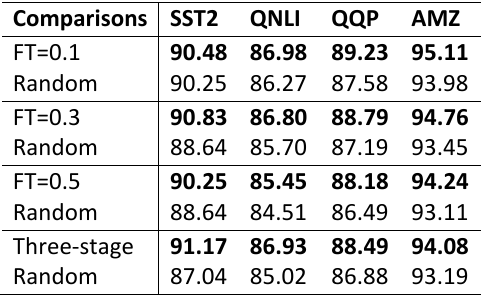}
		\vspace*{-1mm}
	\caption{Using loss thresholds for skipping data is superior to skipping  the same amount of data but is randomly picked. FT means using a fixed threshold; Three-stage uses an automatic threshold. Numbers are accuracy.}
	\label{tab:cmp-random}
		\vspace{-5mm}
\end{table}

%% file: tab-freeze.tex

\begin{table}[t]
\scalebox{0.8}{
\centering
\begin{tabular}{lllll}
\hline
\textbf{Benchmarks}& \textbf{SST2} & \textbf{QNLI} & \textbf{QQP} & \textbf{AMZ} \\
\hline
\textit{TrainAll}+LF &88.99/1.00 &82.17/1.00  &86.64/1.00 &94.29/1.00 \\
Ours+LF &88.53/0.21 &81.15/0.57 &85.03/0.52 &92.79/0.14 \\
\hline
\end{tabular}
}
\vspace{-2mm}
\caption{Accuracy and training time of our method and
\textit{TrainAll} with layer freezing. Training time normalized to \textit{TrainAll}+LF of the same benchmark}
\label{tab:freeze}
\vspace{-5mm}
\end{table}

%% file: summary.tex
\section{Conclusions}
\vspace{-1mm}
We present online data filtering, an efficient training mechanism for optimizing training data usage. We automatically maintain a loss threshold from model losses, then train and query a simple predictor to skip both forward and backward passes. So that unnecessary data instances will be filtered out and we achieve great accuracy-efficiency tradeoff. We formulate two algorithms under the Three-stage training method for three realistic and distinct NLP tasks, sentiment classification, QA/NLI, and paraphrase identification, which leads to consistent improvements over strong baselines.

%% file: ack.tex
\section*{Acknowledgments}
The authors were supported in part by NSF awards \#2128725, \#1919197, \#2106893, \#2124538, and Virginia's Commonwealth Cyber Initiative.
The authors thank the anonymous reviewers for their insightful feedback.